\pdfoutput=1

\documentclass[11pt]{article}

\usepackage[preprint]{acl}

\usepackage{times}
\usepackage{latexsym}
\usepackage{multirow}

\usepackage{amsthm}
\usepackage{amsbsy}
\usepackage{diagbox}
\theoremstyle{definition}

\usepackage[T1]{fontenc}
\usepackage[utf8]{inputenc}
\usepackage{microtype}
\usepackage{inconsolata}
\usepackage{graphicx}
\usepackage{hyperref}
\usepackage{algorithm}
\usepackage{algpseudocode}

\title{Entity Image and Mixed-Modal Image Retrieval Datasets}

\author{
    Cristian-Ioan Blaga$^{1*}$ \quad Paul Suganthan$^{1*\dagger}$ \quad Sahil Dua$^{1*}$ \\ 
    \textbf{Krishna Srinivasan$^{\mathsection}$ \qquad Enrique Alfonseca$^1$ \qquad Peter Dornbach$^1$} \\
    \textbf{Tom Duerig$^1$ \qquad Imed Zitouni$^1$ \qquad Zhe Dong$^{2*\ddagger \mathsection}$} \\
    $^1$Google Switzerland, $^2$Microsoft AI \\
    \small{\textbf{Correspondence:} $^{\dagger}$\href{paulgc@google.com}{paulgc@google.com}, $^{\ddagger}$\href{hoogendong@gmail.com}{hoogendong@gmail.com}.} \\
    \small{$^*$Equal contribution. $^\mathsection$Work done while at Google. }
}

\begin{document}

\maketitle

\begin{abstract}
Despite advances in multimodal learning, challenging benchmarks for mixed-modal image retrieval that combines visual and textual information are lacking. This paper introduces a novel benchmark to rigorously evaluate image retrieval that demands deep cross-modal contextual understanding. We present two new datasets: the Entity Image Dataset (EI), providing canonical images for Wikipedia entities, and the Mixed-Modal Image Retrieval Dataset (MMIR), derived from the WIT dataset. The MMIR benchmark features two challenging query types requiring models to ground textual descriptions in the context of provided visual entities: single entity-image queries (one entity image with descriptive text) and multi-entity-image queries (multiple entity images with relational text). We empirically validate the benchmark's utility as both a training corpus and an evaluation set for mixed-modal retrieval. The quality of both datasets is further affirmed through crowd-sourced human annotations. The datasets are accessible through the GitHub page\footnote{\url{https://github.com/google-research-datasets/wit-retrieval}}.
\end{abstract}

\section{Introduction}
The remarkable progress in multimodal learning has been fueled by the availability of large-scale datasets for multimodal retrieval tasks, such as image-text retrieval tasks MS COCO~\citep{mscoco2014} and Flickr30K~\citep{flickr30k2014}, and text-text retrieval tasks, such as MS MARCO~\citep{msmarco2016} and Natural Questions~\citep{kwiatkowski2019naturalquestions}. However, a comprehensive and challenging benchmark specifically designed for \textbf{mixed-modal image retrieval}, where the query context contains information from both visual and textual modalities, is notably lacking. This paper addresses this critical gap by introducing a novel benchmark designed to rigorously evaluate cross-modal image retrieval with a strong emphasis on understanding contextual information derived from both image and text modalities.

To the best of our knowledge, existing benchmarks do not adequately evaluate a model's ability to retrieve images based on complex queries that combine visual inputs with nuanced textual descriptions. While the Fashion200K dataset~\citep{fashion200k2017} offers a valuable resource, its domain-specific focus (fashion) and relatively limited scale (202K examples) constrain its broader applicability. This absence of a suitable benchmark impedes the thorough evaluation and direct comparison of machine learning models in scenarios demanding a deep understanding of textual context in relation to visual entities within a single query.

To overcome these limitations and foster advancements in this area, we present a new benchmark comprising two distinct datasets: the \textbf{Entity Image Dataset (EI)} and the \textbf{Mixed-Modal Image Retrieval Dataset (MMRI)}. This benchmark is specifically designed to facilitate rigorous evaluation and comparison of diverse machine learning models for mixed-modal understanding and retrieval.

\begin{figure*}[ht]
\centering
\includegraphics[width=.7\textwidth]{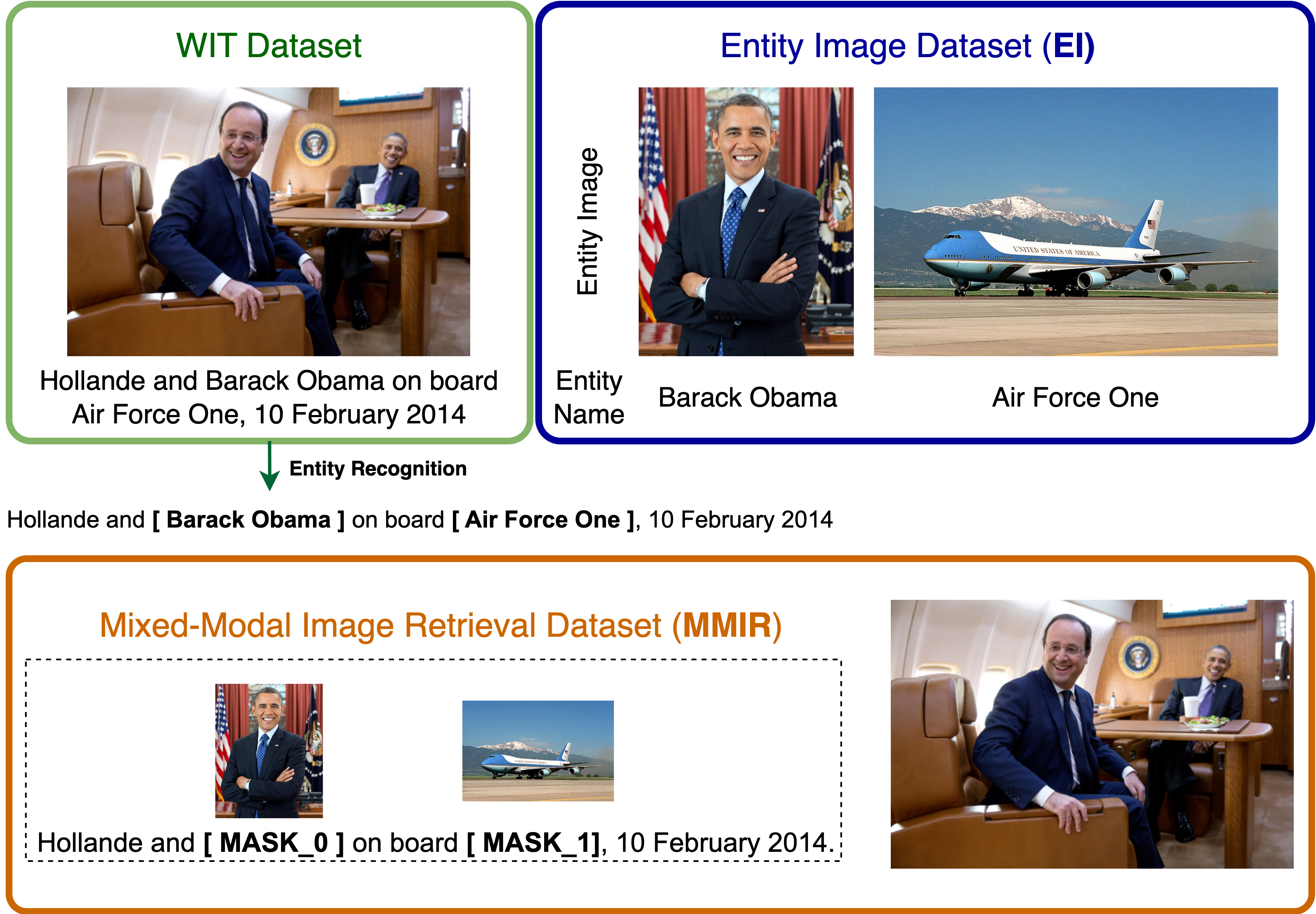}
\captionsetup{width=0.9\textwidth,font=scriptsize,justification=centering}
\caption{Examples of Entity Image Dataset (\textbf{EI}) and Mixed-Modal Image Retrieval Dataset (\textbf{MMIR}). Starting from WIT~\citep{srinivasan2021wit} dataset, \textit{entity recognition} is applied to identify the entities mentioned in the reference description. For each entity, a unique image is identified from Wikipedia, as the \textit{canonical} entity image, in EI dataset. In MMIR dataset, we remove the entity name from original reference description, and replace them with mask ids, \texttt{[MASK\_*]}, which point to the corresponding entity images. The mixed-modal context with masked reference description and referenced entity images will be used to match the original image in the WIT dataset.} 
\label{fig:dataset_example}
\end{figure*}

A defining characteristic of our benchmark is its emphasis on robust contextual understanding, achieved through the inclusion of non-trivial text queries that interact with provided visual inputs, with examples shown in Figure~\ref{fig:dataset_example}. Specifically, the Mixed-Modal Image Retrieval dataset supports two novel query types:

\paragraph{Single entity-image queries.} These queries combine a representative image of a specific entity with a textual description detailing its activity or state, requiring the retrieval of a matching image depicting that entity in the described context.

\paragraph{Multi entity-images queries.} These queries leverage representative images of multiple entities alongside a detailed textual description that articulates the contextual relationships among these entities, demanding the retrieval of an image that accurately portrays the described interaction or scene.

The development of this benchmark, encompassing the meticulous creation of the Entity Image Dataset leveraging Wikipedia data and the targeted annotation of the WIT dataset for our Mixed-Modal Image Retrieval Dataset, signifies a substantial contribution to the field of multimodal image understanding. By providing a challenging and contextually rich evaluation platform, this benchmark has the potential to significantly drive the development of more sophisticated and effective models capable of retrieving images based on intricate mixed-modal inputs. We anticipate that the public release of this benchmark will serve as a valuable and standardized resource for the research community, enabling objective evaluation and comparison of innovative approaches in this rapidly advancing domain.

\section{Methods}

This section introduces the method to build two datasets: the \textit{Entity Image Dataset} (\textbf{EI}), a collection of canonical entity images from Wikimedia, and the \textit{Mixed-Modal Image Retrieval Dataset} (\textbf{MMIR}). MMIR is a benchmark for evaluating models that retrieve images by jointly encoding entity images and their contextual relationship descriptions. Figure~\ref{fig:dataset_example} shows examples from both.

\subsection{Entity Image Dataset}
The Entity Image Dataset (EI) comprises curated canonical images from Wikimedia Commons. For each entity, a single, representative canonical image was extracted from its Wikipedia page to serve as a visual identifier.

Given multiple images and language variations for each Wikipedia entity, a systematic four-stage process was developed to select the most appropriate canonical image: Wikipedia content page crawling, candidate image identification, consolidation, and final selection. This subsection details the methodology and trade-offs for constructing EI.

\subsubsection*{Wikipedia Content Page Crawl}

The initial stage involved crawling all Wikipedia Content\footnote{For example, English content pages can be found in https://en.wikipedia.org/wiki/Wikipedia:Contents.} pages, excluding discussion and comment pages, totaling approximately $124$ million pages across $279$ languages. A Flume~\citep{flume} pipeline was employed for programmatic processing, filtering, cleaning, and storage of the crawled Wikipedia data, following the methodology used in WIT~\citep{srinivasan2021wit}.

\begin{figure}[t]
\centering
\includegraphics[width=.45\textwidth]{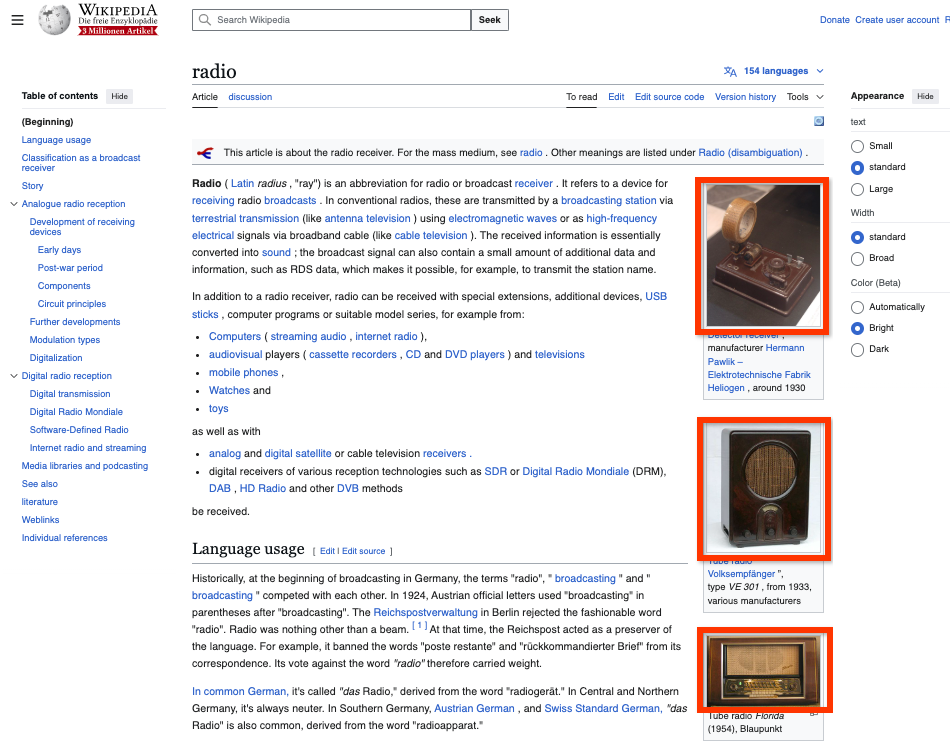}
\captionsetup{width=0.9\textwidth,font=scriptsize,justification=centering}
\caption{An example Wikipedia page with multiple candidate images.} 
\label{fig:wiki_page_candidates}
\end{figure}

\subsubsection*{Candidate Image Identification}
This stage focuses on identifying candidate images for each entity associated with a Wikipedia content page.  An entity represents a real-world object or concept, such as a person, place, organization, creative work, or abstract idea. Pages without an associated entity were excluded from further processing. Figure~\ref{fig:wiki_page_candidates} shows an example Wikipedia content page with multiple candidate images for the entity "radio".

Given a page $p$ with an associated entity $e$, the goal is to identify a set of high-quality candidate images representing $e$. Several criteria were established to ensure image quality and research use:

\begin{itemize}
    \item \textbf{Minimum Resolution:} Images were required to have both a height and width of at least 100 pixels. This ensures a minimum level of detail and avoids very small, low-quality images.
    \item \textbf{Research-Permissive Licensing:} Only images with licenses explicitly permitting research use, such as those under Creative Commons, were considered. This aligns with the CC-BY-SA license governing Wikipedia's textual content and ensures the dataset can be freely used for research purposes.
\end{itemize}

Let $I_p$ represent the set of images from page $p$ meeting the quality and licensing criteria.  We employ an image annotation service, such as \citet{imageannotation}, that takes an image $i \in I_p$ as input and returns predicted entities with confidence scores.  Image $i$ is a candidate for entity $e$ \textit{if and only if} the service predicts $e$ in $i$.  If no candidate images are found for entity $e$ on page $p$, the page is discarded.  Otherwise, the entity $e$ and its associated candidate images from page $p$ are recorded.

To facilitate subsequent processing, we record metadata for each entity and its candidate images. Entity metadata includes the entity type and names, in available languages. Candidate image metadata comprises the image URL, image type, and the section of page $p$ where the image appeared.

\subsubsection*{Candidate Image Consolidation}

The candidate image identification process can result in duplicate image candidates for an entity, and a single entity may have multiple associated Wikipedia pages, due to several factors. These include the image appearing on different language versions of the same entity's Wikipedia page, the image being present in multiple sections of a single page, and the entity being associated with several relevant pages (for example, the entity "Barack Obama" might be linked to both the page of "Barack Obama" and the page of "List of Presidents of the United States").

To address this, we perform a consolidation step.  First, \textit{all} candidate images for a given entity are merged, regardless of which Wikipedia page they originated from. Then, within this merged set, candidate images are grouped based on their unique image URL.

If the same image URL appears multiple times (across different pages or sections), we prioritize the instance that appears \textit{topmost} on its respective Wikipedia page.  This heuristic assumes that images placed higher on a page are more likely to be representative of the entity.

The metadata associated with all instances of the image (page sections, entity names) are aggregated. This provides a comprehensive view of the context.  After consolidation, the dataset comprises over $\sim4.34$M unique entities with candidate images.

\subsubsection*{Canonical Image Selection}
Choosing a representative canonical image from the candidate set for each entity requires a robust selection strategy.  Simple heuristics, such as selecting the topmost image on a Wikipedia page, is proven to be inadequate, particularly when dealing with entities associated with multiple pages (e.g., different language versions).

Our approach combines the \textit{confidence score}, provided by the image annotation service, with the \textit{image's position}, within its source Wikipedia page. Let $C_e = \{c_1, ..., c_n\}$ represent the set of candidate images for entity $e$, and let $s_i$ be the confidence score associated with candidate image $c_i$.

We first apply a confidence threshold, $t$, to filter the candidate set.  A new set, $D_e$, is created, containing only those candidates whose predicted entity matches $e$ with a confidence score greater than or equal to $t$, such that $D_e = \{c_i \in C_e \mid s_i \geq t\}$.

If $D_e$ is empty after this filtering step, the entity $e$ is discarded, as no sufficiently confident candidate image was found.

If $D_e$ is non-empty, the images within $D_e$ are sorted to determine the canonical image by their positions of the sections on the Wikipedia pages containing the images. Images appearing in earlier sections are prioritized, reflecting the assumption that more representative images tend to appear earlier in an article.  If multiple images appear within the same section, the tie is broken using the confidence score $s_i$, with higher confidence scores being prioritized. The image with the highest rank after this sorting process is selected as the canonical image for entity $e$.

The \textit{selection} of the confidence threshold, $t$, significantly impacts the number and quality of entities with canonical images. An initial approach involved randomly sampling entities, having human raters label their candidate images as valid or invalid, and then averaging the optimal thresholds determined for each entity. However, this method is proven to be too aggressive, as it is reducing the number of entities with canonical images to approximately $862$K and disproportionately filtering out key entity types of interest, such as locations and people. Therefore, we adopted a more refined strategy: identifying \textit{key entity types} and determining a \textit{separate}, \textit{optimized} threshold for each type.

\begin{figure*}[ht]
\centering
\includegraphics[width=.7\textwidth]{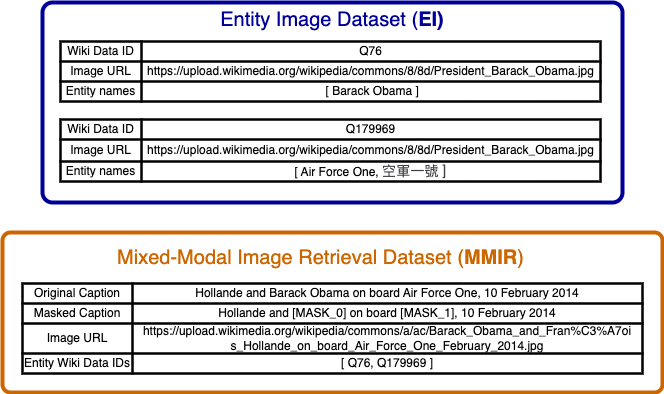}
\captionsetup{width=0.9\textwidth,font=scriptsize,justification=centering}
\caption{Example features of Entity Image Dataset (\textbf{EI}) and Mixed-Modal Image Retrieval Dataset (\textbf{MMIR}). The EI dataset consists of the Wiki data ID for the entity along with the entity names (in available languages) and image information. The MMIR dataset consists of the original caption (from the WIT dataset), the masked caption, the image information (from the WIT dataset), and the Wiki data IDs for the masked entities.} 
\label{fig:example_features}
\end{figure*}

\paragraph{Entities of Interest:} This work focuses on \textit{physical entities}, the entities that can be visually represented, such as people, objects, animals, and locations.  Our goal is to identify high-quality canonical images for these entities.  Specifically, we consider the following $\mathbf{8}$ categories as physical entities: \textit{people}, \textit{animals}, \textit{plants}, \textit{objects}, \textit{locations}, \textit{tourist attractions}, \textit{historical places}, and \textit{historical events}. All remaining entities are grouped into an \textit{other} category. The entity types are annotated by the image annotation services~\citep{imageannotation}.

To address the limitations of the single-threshold approach, we implemented a modified strategy that utilizes \textit{category-specific thresholds}.  For each of the defined entity categories, we independently sampled a set of entities and determined the optimal threshold that maximized agreement with human annotations (as described previously). This category-specific approach significantly improved the coverage, resulting in over $1.79$M entities with canonical images, which is a more than twofold increase compared to the initial approach.

\paragraph{Handling the \textit{Other} Category:} For entities in the \textit{other} category, we aimed to filter out non-physical entities, such as abstract concepts like \textit{Trade} or \textit{Language}, which lack representative canonical images.  During human annotation of candidate images for these entities, raters were given three options: (a) "Entity is non-physical", (b) "Entity is physical, image is not canonical", and (c) "Entity is physical, image is canonical". The threshold for this category was then chosen to minimize the inclusion of non-physical entities.

Further details regarding the Entity Image Dataset can be found in Section~\ref{sec:ei_dataset_info}.

\subsection{Mixed-Modal Image Retrieval Dataset}
\label{sec:method_mmir}

We introduce the Mixed-Modal Image Retrieval (MMIR) dataset, which we construct by leveraging the comprehensive Wikipedia Image Text (WIT) dataset~\citep{srinivasan2021wit}. WIT is a carefully curated dataset of $37$ million image-text pairings, featuring $11$ million distinct images and covering more than $100$ languages. The creation of the MMIR dataset involves two primary steps: filtering and masking.

\subsubsection{Filtering}
The WIT dataset includes various associated texts for each image, such as the reference description, attribution information, and \texttt{alt-text}\footnote{https://en.wikipedia.org/wiki/Alt\_attribute.} of the image. Recognizing that the reference description typically provides the most salient information about the image content, we exclusively considered image-text pairs where the text corresponds to this reference description. This initial filtering step resulted in a refined subset of approximately $17.2$ million examples.

\subsubsection{Masking}
Following the filtering stage, we processed the resulting WIT subset to generate the MMIR examples. For each image-text pair $(i_i, t_i)$, where $i_i$ represents the image and $t_i$ the text, we employed both an image annotation service (such as~\cite{imageannotation}) and a text annotation service (such as~\cite{textannotation}). The image annotation service identified entities present in the image $i_i$ (denoted as the set $I_i$), while the text annotation service identified entities within the text $t_i$ (denoted as the set $T_i$). We then determined the intersection of these entity sets, $P_i = I_i \cap T_i$, representing the common entities identified in both the image and the text. Subsequently, we refined $P_i$ by removing any entities that lacked a corresponding canonical image within the EI dataset. Finally, for each remaining entity in $P_i$, we masked its name within the text description $t_i$, thus generating a mixed-modal image retrieval example as shown in Figure~\ref{fig:example_features}. We discarded any original WIT example for which the set of common, canonical-image-linked entities, $P_i = \emptyset$, was empty.

\section{Data Information}

The datasets are released under CC-BY-SA license, similar to the Wikipedia~\cite{wikilicense}  and WIT~\cite{witlicense} datasets, and can be accessed from the GitHub page: \url{https://github.com/google-research-datasets/wit-retrieval}.

\subsection{The Entity Image Dataset}
\label{sec:ei_dataset_info}
The Entity Image (EI) dataset comprises $1.80$M entities, each associated with a canonical image. For each entity, the dataset includes its Wikidata ID, entity names (in various available languages), and comprehensive image information such as the image URL and type. Figure~\ref{fig:example_features} illustrates sample entries from the EI dataset for the entities \textit{Barack Obama} and \textit{Air Force One}.

Table~\ref{tab:entity-type-stats} presents the distribution of examples across different entity categories within the EI dataset. As the table reveals, the dataset is predominantly composed of entities belonging to the \textit{Person} and place-related categories (\textit{Locations}, \textit{Tourist Attraction}, \textit{Historical Places}).

\subsection{Mixed-Modal Image Retrieval Dataset}

The Mixed-Modal Image Retrieval (MMIR) dataset encompasses over $9$M examples spanning more than $100$ languages, partitioned into \textit{train}, \textit{validation}, and \textit{test} splits. To ensure consistency, we maintained the original data splits from the WIT dataset; specifically, each of the MMIR splits was derived from the corresponding WIT split through the entity annotation process, as detailed in \ref{sec:method_mmir}. Detailed statistics for each split are presented in Table~\ref{tab:mmir-data-stats}. To the best of our knowledge, MMIR is the largest multilingual mixed-modal retrieval dataset currently available.

Statistics on the number of entities per example in the MMIR dataset are detailed in Table~\ref{tab:mmir-entity-stats}. Notably, roughly 50\% of the examples include only one entity, and more than 90\% of the dataset contains a maximum of three entities.

\begin{table}
\centering
\footnotesize
\begin{tabular}{l r}
  \hline
  \hline
  Entity Type & Count \\ 
  \hline
  Person & 775.18K \\ 
  Animal & 40.63K  \\
  Plants & 13.06K  \\
  Locations & 222.72K \\
  Tourist Attraction & 189.26K \\
  Historical Places & 43.76K \\
  Historical Events & 25.95K \\
  Objects & 1.63K \\
  Other &  486.14K \\
  \hline
  Total & $1.80$M \\
  \hline
\end{tabular}
\captionsetup{width=0.9\linewidth,font=footnotesize,justification=centering}
\caption{Number of examples per entity type in the Entity Image (EI) dataset.}
\label{tab:entity-type-stats}
\end{table}

\begin{table}
\centering
\footnotesize
\begin{tabular}{c c c c}
  \hline
  Type & Train & Validation & Test \\ 
  \hline
  Examples & $9.06$M & $46.33$K & $56.67$K \\ 
  Languages & $108$ & $107$ & $108$ \\
  Avg. word tokens & $8.5$ & $8.4$ & $8.7$ \\
  \hline
\end{tabular}
\captionsetup{width=0.9\linewidth,font=footnotesize,justification=centering}
\caption{Statistics for each split of the MMIR dataset.}
\label{tab:mmir-data-stats}
\end{table}

\begin{table}
\footnotesize
\centering
\begin{tabular}{c r}
  \hline
  \hline
  \# Entities &  Count \\
  \hline
  1 & 4.79M \\ 
  2 & 2.54M  \\
  3 & 1.05M  \\
  4 & 425.37K \\
  5 & 181.29K \\
  > 5 & 171.18K \\
  \hline
\end{tabular}
\captionsetup{width=0.9\linewidth,font=footnotesize,justification=centering}
\caption{Statistics on the number of entities per example in the MMIR dataset.}
\label{tab:mmir-entity-stats}
\end{table}

\section{Model-based Evaluation of MMIR}

This section assesses the Mixed-Modal Image Retrieval (MMIR) dataset's utility as both a training corpus and an evaluation benchmark for cross-modal retrieval, using model-based evaluations.

\subsection{Experimental Setup}

\begin{figure}[h]
\centering
\includegraphics[width=0.95\linewidth]{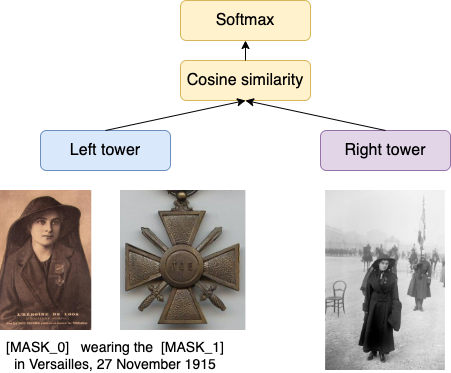}
\captionsetup{width=0.95\linewidth,font=footnotesize,justification=centering}
\caption{Shared-parameter dual-encoder architecture for MMIR model evaluation.} 
\label{fig:de_arch}
\end{figure}

We employed a dual-encoder architecture with shared parameters, inspired by~\citet{dong2022exploringdualencoderarchitectures}. We used a visual-language model architecture similar to PaLI~\citep{chen2023pali}, to process the mixed-modal information from visual and textual inputs. The text encoder was initialized with mT5-Base~\citep{xue2021mt5massivelymultilingualpretrained}, and the vision encoder with ViT-Large~\citep{dosovitskiy2021imageworth16x16words}. The shared encoder processed both visual and textual embeddings, which were then mean-pooled, linearly projected to 768 dimensions, and optimized using in-batch sampled softmax loss based on cosine similarity, as shwon in Figure~\ref{fig:de_arch}.

We fine-tuned the WebLI-10B pre-trained model~\citep{chen2023pali} under three settings: (a) $14$ epochs on MMIR, (b) $7$ epochs on CC3M~\citep{changpinyo2021conceptual12mpushingwebscale}, and (c) $14$ epochs on MMIR followed by $7$ epochs on CC3M. CC3M is a large-scale dataset of approximately $3$ million image-caption pairs sourced from web alt-text, providing diverse and realistic image-text relationships for training multimodal models. We then compared the performance of these fine-tuned models with the zero-shot performance of the pre-trained model.

For fine-tuning, we used the Adafactor optimizer~\citep{shazeer2018adafactoradaptivelearningrates} with batch size $1024$ and learning rate $10^{-3}$. The vision encoder was frozen during the training. Given that the majority of MMIR examples contain 5 or fewer entities, as shown in Table~\ref{tab:mmir-entity-stats}, we filtered the MMIR training set by removing examples with more than 5 entities.

\subsection{Evaluation Results}

We evaluated the fine-tuned models on three benchmarks with Recall@K, K$=1, 5, 10$: the MMIR evaluation set, and the established Flickr30k~\citep{flickr30k2014}, and MS-COCO~\citep{mscoco2014} benchmarks. This multi-faceted evaluation aimed to: (1) validate MMIR as a novel and challenging evaluation benchmark for mixed-modal / cross-modal retrieval, (2) quantify the impact of incorporating MMIR training data on model performance, both on MMIR and on generalizability to Flickr30k and MS-COCO, and (3) assess model robustness across diverse evaluation scenarios.

\begin{table*}
\centering
\newcommand\topstruct{\rule{0pt}{2.2ex}} 
\begin{tabular}{cc||cc|cc|cc|cc}
    \hline 
    \multicolumn{2}{c||}{\multirow{2}{*}{\diagbox{Eval Set}{Training Set}}}
    & \multicolumn{2}{c|}{0-shot} 
    & \multicolumn{2}{c|}{MMIR} 
    & \multicolumn{2}{c|}{CC3M}
    & \multicolumn{2}{c}{Combined} \topstruct \\
    & & R@1 & R@5 & R@1 & R@5 & R@1 & R@5 & R@1 & R@5 
    \topstruct \\
    \hline \hline
    \multirow{2}{*}{MMIR} 
    & $\mathrm{I}+\mathrm{T}\,\to\,\mathrm{I}$
    & $7.32$ & $25.22$ 
    & $\mathbf{13.17}$ & $\mathbf{41.23}$
    & $5.83$ & $20.65$ 
    & $12.19$ & $40.29$ \topstruct \\
    & $\mathrm{I}\,\to\,\mathrm{I}+\mathrm{T}$
    & $8.10$ & $26.17$
    & $\mathbf{13.36}$ & $\mathbf{40.51}$ 
    & $6.45$ & $22.51$ 
    & $11.90$ & $39.27$ \topstruct \\
    \hline \hline
    \multirow{2}{*}{Flickr30k}  
    & $\mathrm{I}\,\to\,\mathrm{T}$
    & $63.60$ & $93.00$ 
    & $48.90$ & $88.10$
    & $\mathbf{65.10}$ & $94.00$
    & $64.90$ & $\mathbf{94.10}$ \topstruct \\
    & $\mathrm{T}\,\to\,\mathrm{I}$
    & $44.98$ & $81.72$
    & $34.46$ & $74.12$ 
    & $51.10$ & $\mathbf{86.54}$  
    & $\mathbf{52.04}$ & $86.32$ \topstruct \\
    \hline
    \multirow{2}{*}{MS COCO}  
    & $\mathrm{I}\,\to\,\mathrm{T}$
    & $30.34$ & $67.46$
    & $22.18$ & $58.02$
    & $\mathbf{36.56}$ & $73.54$
    & $36.44$ & $\mathbf{73.92}$ \topstruct \\
    & $\mathrm{T}\,\to\,\mathrm{I}$
    & $21.84$ & $55.04$
    & $10.28$ & $39.06$
    & $27.56$ & $\pmb{63.47}$
    & $\pmb{27.75}$ & $63.46$ \topstruct \\
    \hline
\end{tabular}
\captionsetup{width=0.9\linewidth,font=small,justification=centering}
\caption{Model-based retrieval evaluation on MMIR, Flickr30k, and MS COCO, comparing zero-shot and the models fine-tuned with MMIR, CC3M, MMIR+CC3M. Best Recall@1/5 are highlighted. Retrieval tasks: Image-to-Text (I$\to$T) and Text-to-Image (T$\to$I) for Flickr30k and MS COCO, Image+Text-to-Image (I+T$\to$I) and Image-to-Image+Text (I$\to$I+T) for MMIR.}
\label{tab:model-eval}
\end{table*}

\begin{table}
\centering
\footnotesize
\renewcommand{\arraystretch}{1.3}
\begin{tabular}{c c}
  \hline
  Majority Rating & Percentage \\
  \hline
  Excellent & 76.8\% \\
  Good & 20.5\% \\
  Average & 1.5\% \\
  Poor & 0.9\% \\
  Very Poor & 0.3\% \\
  \hline
\end{tabular}
\captionsetup{width=0.9\linewidth,font=footnotesize,justification=centering}
\caption{Distribution of majority ratings for how well canonical image matches the entity in the EI dataset. Majority rating is computed as the rating with the highest consensus among the raters.}
\label{tab:eval-ei-matching}
\end{table}

\begin{table*}[h]
\centering
\footnotesize
\renewcommand{\arraystretch}{1.5}
\begin{tabular}{l r r r r r r r r}
  \hline
  Majority Rate & Plants & Hist. Events & Locations & Person & Tour. Attr. & Animal & Hist. Sites & Objects \\
  \hline \hline
  Excellent & 72.13\% & 55.46\% & 52.69\% & 92.0\% & 73.97\% & 81.56\% & 83.14\% & 75.54\% \\
  Good & 23.19\% & 35.26\% & 44.29\% & 7.38\% & 25.08\% & 15.92\% & 15.29\% & 21.94\% \\
  Average & 1.91\% & 7.22\% & 1.29\% & 0.62\% & 0.32\% & 1.55\% & 1.57\% & 1.94\% \\
  Poor & 1.91\% & 1.65\% & 1.51\% & 0.0\% & 0.63\% & 0.39\% & 0.0\% & 0.58\% \\
  Very Poor & 0.86\% & 0.41\% & 0.22\% & 0.0\% & 0.0\% & 0.58\% & 0.0\% & 0.0\% \\
  \hline \hline
\end{tabular}
\captionsetup{width=0.9\linewidth,font=footnotesize,justification=centering}
\caption{Distribution of Majority Ratings per category for how well Canonical Image matches the Entity for the EI dataset. Majority rating is computed as the rating with the highest consensus among the raters.}
\label{tab:eval-ei-matching-by-type}
\end{table*}

\subsubsection{Evaluation on MMIR Eval Set}

We evaluated the MMIR dataset as a retrieval benchmark by comparing various fine-tuned models, including a zero-shot model. As shown in Table~\ref{tab:model-eval}, models fine-tuned on MMIR demonstrate a substantial advantage over those trained on CC3M. This underscores MMIR's complex nature, demanding a cohesive understanding of both image and text – a necessity amplified by multiple entity images and mixed textual information, in contrast to CC3M's single-modality input per encoder tower. The strong performance of the model fine-tuned with both MMIR and CC3M further validates MMIR's value as a supplementary dataset that enhances the mixed-modal comprehension abilities lacking in CC3M.

\subsubsection{Evaluations on Flickr30k \& MS COCO Eval Sets}

The preceding section demonstrated the efficacy of the MMIR dataset in enhancing mixed-modal understanding capabilities. Building on this, we now evaluate the impact of incorporating MMIR into training on the standard image-text retrieval benchmarks, namely Flickr30k~\citep{flickr30k2014} and MS-COCO~\citep{mscoco2014}. This evaluation compares the retrieval metrics of various fine-tuned models, including the zero-shot pre-trained model. As shown in last $4$ rows of Table~\ref{tab:model-eval}, the model fine-tuned on both the MMIR and CC3M datasets performs at a level comparable to the model fine-tuned solely on CC3M. This indicates that integrating the MMIR training data into the existing fine-tuning strategy does not negatively impact model performance on the standard Flickr30k and MS-COCO benchmarks.

\section{Human Evaluation}

While model-based evaluation demonstrated the MMIR dataset's utility in improving mixed-modal understanding and its robustness as a training set, human evaluation provides a crucial validation of the dataset's quality. In this section, we employed crowd-sourced annotators to assess the quality of both the Entity Image (EI) and Mixed-Modal Image Retrieval (MMIR) datasets. For EI, we aimed to validate the representativeness of the selected canonical images for their respective entities. For MMIR, we evaluated the semantic coherence between the masked caption and entity images in relation to the original caption.

For annotation, we employed a fixed human rater pool based in US/Canada and India. The raters were compensated appropriately taking their location and the complexity of the task into account. The raters were given detailed instructions about the tasks as shown in the evaluation templates in Figure~\ref{fig:human_eval_template}.

\subsection{Evaluating the EI Dataset}

\begin{figure*}[t]
\centering
\includegraphics[width=.95\linewidth]{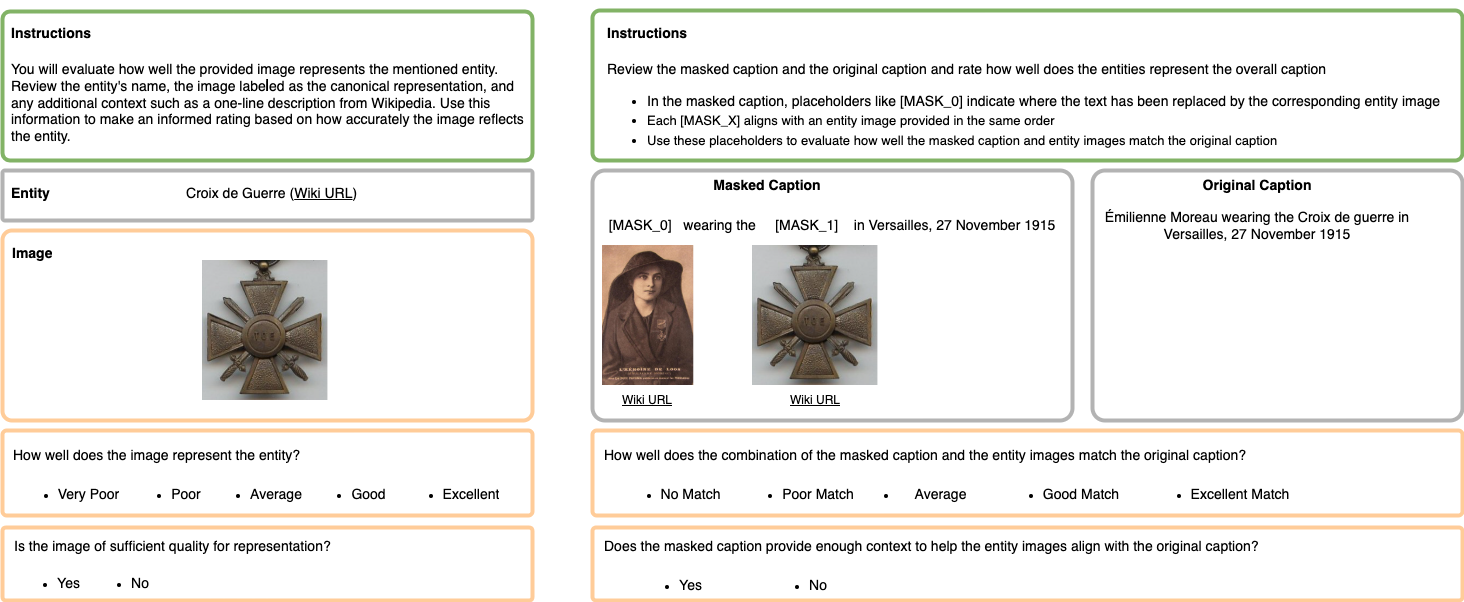}
\captionsetup{width=0.9\linewidth,font=scriptsize,justification=centering}
\caption{Human evaluation template for \textbf{EI} dataset (on the left) and \textbf{MMIR} dataset (on the right).} 
\label{fig:human_eval_template}
\end{figure*}

We randomly sampled $680$ English examples from the EI dataset, with each example evaluated by $5$ independent human raters. Raters assessed how well the provided canonical image represents the corresponding entity, guided by the evaluation template shown on the left in Figure~\ref{fig:human_eval_template}.

As shown in Table~\ref{tab:eval-ei-matching}, the majority ratings for the canonical image's match to the entity were overwhelmingly positive, with over $97\%$ of examples rated \textit{Good} or \textit{Excellent}. The \textit{majority rating} was determined by the highest consensus among the $5$ raters. This strong agreement underscores the high quality of the canonical image selection process in the EI dataset, indicating that the chosen images effectively represent their associated entities.

The per-category distribution of these majority ratings, detailed in Table~\ref{tab:eval-ei-matching-by-type}, further demonstrates the consistency of this high quality. Across all entity categories, at least $90\%$ of the examples received ratings of \textit{Good} or \textit{Excellent}, suggesting a robust and effective canonical image selection process across diverse entity types within the EI dataset.

In addition to evaluating the semantic match between the entity and its canonical image, we also tasked the human raters with directly assessing the visual quality of these images. The results indicated that for over $99\%$ of the examples, the canonical image was judged to be of good quality.

\subsection{Evaluating the MMIR Dataset}

Given that prior work~\cite{srinivasan2021wit} has already validated the quality and image relevance of the original descriptions within the WIT dataset (the source of our captions), our human evaluation for the MMIR dataset focused on a different critical aspect. Specifically, we assessed whether the masked caption, when presented alongside the corresponding entity image, provides sufficient contextual information to maintain semantic coherence with the original, unmasked caption.

For this evaluation, we randomly sampled approximately $2350$ English examples from the MMIR dataset, with each example evaluated by $5$ independent human raters (guided by the evaluation template shown on the right in Figure~\ref{fig:human_eval_template}). The majority ratings for the semantic coherence between the masked caption, the entity image, and the original caption were positive, with over $81$\% of examples rated \textit{Good} or \textit{Excellent}. The \textit{majority rating} was determined by the highest consensus among the $5$ raters. This strong result underscores the high quality of MMIR dataset's creation, indicating that the masked caption, in conjunction with the entity images, successfully provides the necessary context to align with the meaning of the original caption.

\begin{table}
\centering
\footnotesize
\renewcommand{\arraystretch}{1.3}
\begin{tabular}{l | r r r r}
    \hline
    \hline
    \multirow{2}{*}{Major Rate} & \multicolumn{4}{c}{\# Entities} \\
    & 1 & 2 & 3 & 3+ \\
    \hline
    Excellent & 55.3\% & 62.5\% & 82.1\% & 68.8\% \\
    Good & 18.2\% & 15.9\% & 11.0\% & 25.0\% \\
    Average & 5.0\% & 8.4\% & 3.6\% & 3.9\% \\
    Poor & 8.3\% & 9.2\% & 0.5\% & 0.5\% \\
    Very Poor & 9.9\% & 1.9\% & 0.0\% & 0.0\% \\
    Undecided & 3.2\% & 2.1\% & 2.8\% & 1.8\% \\
    \hline
\end{tabular}
\captionsetup{width=0.9\linewidth,font=footnotesize,justification=centering}
\caption{Distribution of Majority Ratings for how well the combination of entity images and masked caption matches the original full caption for the MMIR dataset.}
\label{human-eval-mmir-entity-image-match-majority}
\end{table}

\section{Conclusion}
Our work proposed a new benchmark for evaluating mixed-modal image retrieval. Concretely, we presented two new datasets: the \textbf{Entity Image Dataset}, providing canonical entity images, and the \textbf{Mixed-Modal Image Retrieval Dataset}, derived from the WIT dataset. We empirically validate the dataset's utility as a training corpus and evaluation set for mixed-modal retrieval. Furthermore, the quality of both datasets was affirmed through crowd-sourced human annotations.

\section{Limitations}
While the EI and MMIR datasets have been generated with a snapshot of Wikipedia data, there are new entities and images being added in Wikipedia and facts being evolved over time. For example, over time some of the facts (captured in captions) may become stale or images being changed or deleted. Hence, it is important to regenerate the datasets with updated Wikipedia snapshots over time to create a more relevant corpus. 

\bibliography{main}

\end{document}